\documentclass[letterpaper, 10 pt, conference]{ieeeconf}  

\IEEEoverridecommandlockouts                              
\usepackage{graphics} 
\usepackage{amsthm, amsmath, amssymb}
\usepackage{listings}
\usepackage{relsize}
\usepackage{mathtools}
\newcommand{\R}{\mathbb{R}}
\newcommand{\overbar}[1]{\mkern 1.5mu\overline{\mkern-1.5mu#1\mkern-1.5mu}\mkern 1.5mu}
\usepackage{dsfont}
\usepackage{hhline}
\usepackage{csquotes}
\usepackage{siunitx}
\usepackage{algorithm}
\usepackage{algorithmic}
\usepackage{booktabs}
\numberwithin{equation}{section}
\usepackage{multirow}
\usepackage{dsfont}

\usepackage{subcaption}
\usepackage{textgreek}
\DeclareMathOperator*{\argmin}{argmin} 

\newcommand{\df}[1]{\ensuremath{\operatorname{d}\!{#1}}}

\numberwithin{equation}{section}

\newtheorem{theorem}{Theorem}

\title{\LARGE \bf Optimal Control of Probabilistic Dynamics Models \\ via Mean Hamiltonian Minimization}

\author{David Leeftink$^{1}$, Çağatay Yıldız$^{2}$, Steffen Ridderbusch$^{3}$, Max Hinne$^{1}$, Marcel van Gerven$^{1}$
\thanks{*This publication is part of the project ROBUST: Trustworthy AI-based Systems for Sustainable Growth with project number KICH3.LTP.20.006, which is (partly) financed by the Dutch Research Council (NWO), ASMPT, and the Dutch Ministry of Economic Affairs and Climate Policy (EZK) under the program LTP KIC 2020-2023. All content represents the opinion of the authors, which is not necessarily shared or endorsed by their respective employers and/or sponsors.}
\thanks{$^{1}$D. Leeftink, M. Hinne and M. van Gerven are with the Department of Machine Learning and Neural Computing, Donders Institute for Brain, Cognition and Behaviour, Radboud University, 6525XZ Nijmegen, the Netherlands. $^{2}$Ç. Yıldız is with the Cluster of Excellence Machine Learning, University of Tübingen, 72076 Tübingen, Germany, and funded by the Deutsche Forschungsgemeinschaft (DFG, German Research Foundation) under Germany’s Excellence Strategy – EXC number 2064/1 $^{3}$S. Ridderbusch is with the Department of Engineering, Control Group, University of Oxford, OX1 3PJ Oxford, UK. (Contact: \texttt{david.leeftink@ru.nl})} %
}

\usepackage{fancyhdr}
\fancypagestyle{firstpage}{%
  \fancyhf{} 
  \fancyfoot[L]{\normalfont \sffamily \scriptsize \lowernote}
}
\fancypagestyle{plain}{%
  \fancyhf{}
  \fancyfoot[L]{\normalfont \sffamily \scriptsize \lowernote}
}

\newcommand{\lowernote}{
\textbf{Accepted final version. \textcopyright\; 2025
IEEE.}
To be presented at the Conference on Decision and Control (CDC), 2025.
Personal use of this material is permitted. Permission from IEEE must be obtained for all other uses, in any current or future media,
including reprinting/republishing this material for advertising or promotional purposes, creating new collective works, for resale or redistribution to servers
or lists, or reuse of any copyrighted component of this work in other works.
}
\begin{document}

\IEEEoverridecommandlockouts

\maketitle
\thispagestyle{firstpage}

\pagestyle{empty}


\begin{abstract}
Without exact knowledge of the true system dynamics, optimal control of non-linear continuous-time systems requires careful treatment under epistemic uncertainty. In this work, we translate a probabilistic interpretation of the Pontryagin maximum principle to the challenge of optimal control with learned probabilistic dynamics models. Our framework provides a principled treatment of epistemic uncertainty by minimizing the mean Hamiltonian with respect to a posterior distribution over the system dynamics.

We propose a multiple shooting numerical method that leverages mean Hamiltonian minimization and is scalable to large-scale probabilistic dynamics models, including ensemble neural ordinary differential equations. Comparisons against other baselines in online and offline model-based reinforcement learning tasks show that our probabilistic Hamiltonian approach leads to reduced trial costs in offline settings and achieves competitive performance in online scenarios. By bridging optimal control and reinforcement learning, our approach offers a principled and practical framework for controlling uncertain systems with learned dynamics.
\end{abstract}


\section{Introduction}
Optimal decision-making under uncertainty is a fundamental challenge in learning-based control, especially when the system dynamics are unknown and must be inferred from limited, noisy data. 
In such settings it is crucial to achieve high data-efficiency, as real-world experiments often come with high costs or risks. To address this, one has to effectively handle epistemic uncertainty, arising from incomplete data, which critically shapes both the robustness of the controller and the balance between exploration and exploitation in learning.

A common approach for decision making under uncertainty involves minimizing the mean cost over the posterior distribution of system dynamics. However, this optimization can become challenging when the posterior distribution is multi-modal or exhibits significant variance. While black-box stochastic optimization methods are often employed to avoid local minima, they overlook the inherent structure of the optimal control problem~\cite{powell2011approximate}.

Pontryagin's maximum principle (PMP) offers a compelling alternative. By defining necessary conditions for optimality, PMP indirectly characterizes optimal control inputs, providing a powerful lens into the structure of optimal solutions~\cite{pontryagin1987}. As a cornerstone of optimal control theory, PMP has driven significant advances in deterministic and stochastic control settings~\cite{kirk2004optimal,peng1990general}. Its ability to leverage the inherent structure of the optimal control problem enables computationally efficient optimization and insights into optimal behavior. 
\begin{figure*}[t]
    \centering
    \includegraphics[width=1.\linewidth]{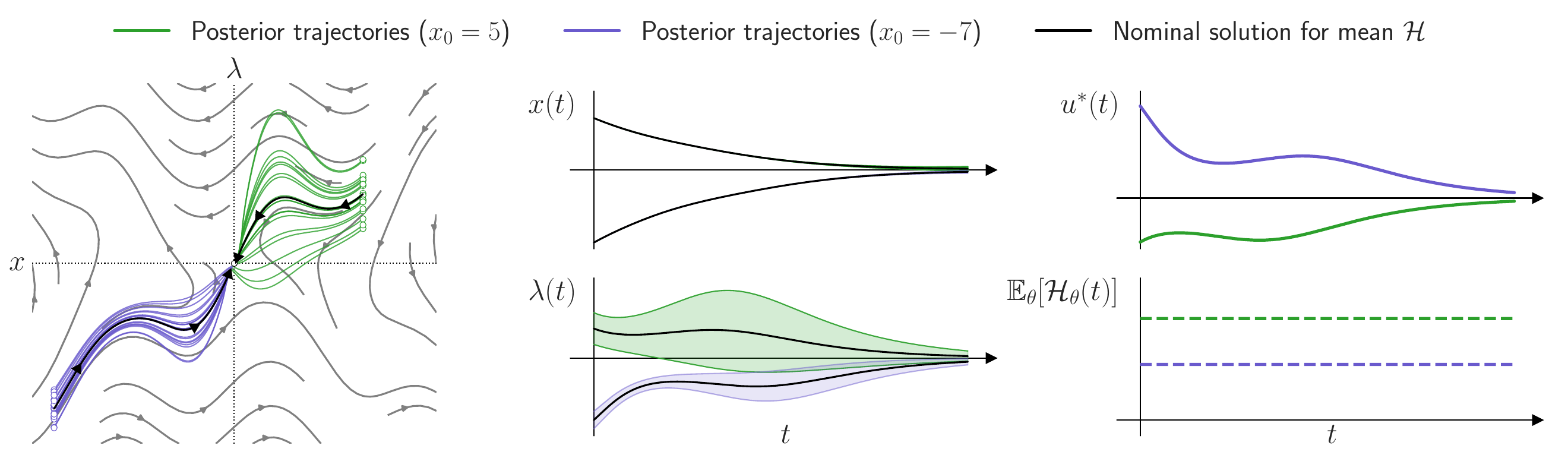}
    \caption{Optimal control of an uncertain dynamical system using the mean Hamiltonian minimization principle. \textbf{(Left)} The Hamiltonian vector field for a system with dynamics depicting state and co-states of the dynamics function $f_{\theta}(x,u)=\theta \sin(x)+u$. The thin lines show posterior trajectories sampled from the dynamics belief $p(\theta)$, while the bold black line shows the nominal solution of the control function that minimizes the mean Hamiltonian. \textbf{(Middle \& Right)} The evolution of the state $x(t)$, co-state $\lambda (t)$, optimal control $u^*(t)$, and the mean Hamiltonian $\mathbb{E}_{\theta}[\mathcal{H}_{\theta}]$. The optimal control function, derived from minimizing the mean Hamiltonian, successfully steers the system to the origin. As a necessary condition for optimality, the mean Hamiltonian is constant under the optimal control.}
    \label{fig:introduction-figure}
\end{figure*}

\textbf{Related work.} Optimal control theory provides foundational principles for learning-based control, guiding optimal action selection under uncertainty. Techniques like the iterative Linear Quadratic Regulator (iLQR)~\cite{li2004iterative} and path integral (PI) control~\cite{kappen2005linear} have been widely adopted to design optimal policies in uncertain or learning-driven settings. Originally developed for deterministic systems, PMP has also been extended to stochastic~\cite{peng1990general}, robust~\cite{bacsar2008h}, and, more recently, learning-based control~\cite{jin2020pontryagin, eberhard2024pontryagin}. While these PMP-based methods primarily address deterministic systems or those with aleatoric uncertainty, recent theoretical advances have established optimality conditions for systems with epistemic uncertainty in their dynamics. In~\cite{bettiol2019averagecost}, conditions are derived for minimizing average terminal costs in general systems with parametric uncertainty, while in~\cite{scagliotti2023optimal} similar conditions are developed for control-affine ensemble systems. Our work builds on this foundation by operationalizing the principle of mean Hamiltonian minimization for learned probabilistic dynamics.  

Our approach closely aligns with Bayesian control and reinforcement learning (RL) paradigms, where optimal policies explicitly account for uncertainty~\cite{ghavamzadeh2015bayesian, pesare2021convergence}. The probabilistic inference for learning and control (PILCO) framework~\cite{deisenroth2011pilco} exemplifies this by learning a Gaussian process (GP) dynamics model and computing policy gradients via closed-form Gaussian approximations. Similarly, Bayesian extensions to model predictive control (MPC) have shown efficacy by planning under uncertain dynamics, using PMP conditions to propagate Gaussian uncertainty over a moving horizon~\cite{kamthe2018data}. 

In deep model-based reinforcement learning (MBRL), methods like probabilistic ensembles with trajectory sampling (PETS)~\cite{chua2018deep} have advanced high-dimensional task solving through uncertainty-aware planning. Our work bridges these advances with continuous-time RL, where discrete-time measurements inform continuous-time dynamics and control~\cite{yildiz2021continuous, treven2023efficient}. By integrating PMP with these Bayesian and continuous-time approaches, we propose a novel framework for optimal planning under epistemic uncertainty, tailored to ensemble-based RL, offering a robust and principled approach to controlling complex, uncertain systems.

\textbf{Contribution.} Our contributions are threefold. First, we formulate the optimal control problem for learned probabilistic dynamics models within the PMP framework, introducing a principled approach to planning based on mean Hamiltonian minimization that explicitly addresses epistemic uncertainty. Second, we develop a practical numerical method combining forward multiple shooting with moving horizon optimization, tailored to probabilistic ordinary differential equation (ODE) models. Third, we validate our approach experimentally, demonstrating improved or competitive performance compared to state-of-the-art methods in both offline and online RL scenarios, highlighting its potential for efficient learning and robust decision-making. 

The application of our framework to a representative optimal control problem with uncertain dynamics is illustrated in Figure~\ref{fig:introduction-figure}, which visualizes the full probabilistic solution, from the state-costate trajectories to the final optimal control.

\section{Problem setting} \label{sec:2}
We study a continuous-time deterministic differential equation $\boldsymbol{f}$, whose solution is given by:
\begin{equation}
    \boldsymbol{x}(t) = \boldsymbol{x}_0 + \int_{t_0}^{t_f} \boldsymbol{f} \left(\boldsymbol{x}(\tau), \boldsymbol{u}(\tau), \tau \right) \df \tau  , \label{eq:trueODE} 
\end{equation}
where $\boldsymbol{x}_0 \in \mathcal{X} \subset \mathds{R}^{d_x}$ is the initial state, $\boldsymbol{u}\colon [t_0, t_f] \to \mathcal{U} \subset \mathds{R}^{d_u}$ is the control input, and $d_x$ and $d_u$ denote the dimensions of the state space and the control input. The objective is to find the optimal control input $\boldsymbol{u}^*$, that is, the input that minimizes the cost function:
\begin{align}
&&C(\boldsymbol{u}) &\coloneq \int_{t_0}^{t_f} L\bigl(\boldsymbol{x}(\tau), \boldsymbol{u}(\tau)\bigr) \df \tau + \Phi \left(\boldsymbol{x}(t_f)\right) ,\label{eq:costfunction}  
\end{align}
where $\Phi(\boldsymbol{x})$ is the terminal cost associated with ending in state $\boldsymbol{x}(t_f)$ at time $t_f$ and $L(\boldsymbol{x},\boldsymbol{u})$ is the state cost associated with taking action $\boldsymbol{u}$ in state $\boldsymbol{x}$. Both the terminal cost and state cost are assumed differentiable with respect to their inputs. 
We are interested in solving the following optimal control problem:
\begin{align}
    \boldsymbol{u}^* (t) &\coloneq \argmin_{\boldsymbol{u}\in \mathcal{U}} C(\boldsymbol{u}) \\
    \text{s.t.}
    \quad \dot{\boldsymbol{x}}(t) &= \boldsymbol{f}\bigl(\boldsymbol{x}(t),\boldsymbol{u}(t), t\bigr) ,\quad \text{for } t \in [t_0, t_f] \enspace \nonumber \\ 
    \boldsymbol{x}(t_0) &= \boldsymbol{x}_0  . \nonumber
\end{align}

Although the true function $\boldsymbol{f}$ is unknown, it can be evaluated by interacting with the system and collecting data over the course of $N$ trials over the time horizon $t_f$. Let $\mathcal{D}_n \coloneq \{ (t^n_i, \boldsymbol{y}^n_i, \boldsymbol{u}^n_i ) \}$ denote the data collected in trial $n\in [1,N]$. Here, $t^n_i \in [t_0, t_f]$ is the time point corresponding to observation $i$ in trial $n$, $\boldsymbol{x}_i^n \coloneq \boldsymbol{x}(t^n_i)$ is the true system state at time $t^n_i$ in trial $n$, $\boldsymbol{y}^n_i \sim \mathcal{N}(\boldsymbol{x}_i^n, \sigma_n^2 \boldsymbol{I}  )$ is a noisy measurement of the state with i.i.d. Gaussian noise with variance $\sigma_n^2$, and $\boldsymbol{u}^n_i$ is the applied control input corresponding to observation $i$ for trial $n$, assumed to be exactly known. It is assumed that the true control function can be reconstructed for $t \in [t_0, t_f]$ using interpolation techniques.

In MBRL, a dynamics model is learned based on previous interactions with the system. In the online variant, this dataset consists initially of only a single trial generated by a random controller. Then iteratively, a model is trained on the available data, the optimal action is determined using the model, and the newly observed trajectory is added to the data.

\subsection{Probabilistic dynamics models}
When learning a dynamics model from limited data, it is crucial to employ efficient exploration strategies that separate epistemic uncertainty, associated with reducible uncertainty due to lack of knowledge of the true system, from aleatoric uncertainty, associated with irreducible uncertainty due to inherent randomness such as sensor or process noise. A general approach for quantifying uncertainty is to assign a prior distribution over the parameters and a likelihood to the possibly noisy observations, and use Bayes' theorem to obtain the posterior distribution over model parameters:
\begin{align}
    \boldsymbol{\theta} &\sim p(\boldsymbol{\theta}), \\  
    p(\mathcal{D} \mid \boldsymbol{\theta}) &= \prod_{i=1}^{T}\mathcal{N}\left(\boldsymbol{y}_i \mid \boldsymbol{x}_{ \boldsymbol{\theta}}(t_i), \sigma^{2} \boldsymbol{I}\right), \\ 
    p(\boldsymbol{\theta} \mid \mathcal{D}) &\propto p(\mathcal{D} \mid \boldsymbol{\theta}) \: p( \boldsymbol{\theta}).  \label{eq:BayesPosterior}
\end{align}
Here, $T$ is the number of discrete-time observations throughout $[t_0, t_f]$ and $\boldsymbol{\theta} \in \Theta = \R^{p}$ are the parameters. For a given realization $\boldsymbol{\theta}$, the following continuous-time model is defined: 
\begin{align}
    \dot{\boldsymbol{x}}_{ \boldsymbol{\theta}}(t) &= \boldsymbol{f}_{ \boldsymbol{\theta}}\bigl(\boldsymbol{x}_{ \boldsymbol{\theta}}(t), \boldsymbol{u}(t), t ) , 
\end{align}
where $\boldsymbol{f}_{ \boldsymbol{\theta}}$ is differentiable in its first two arguments. The posterior distribution can then be estimated using Bayesian sampling methods such as Hamiltonian Monte Carlo (HMC)~\cite{duane1987hybrid}. 

\subsection{Mean cost objective}
A common objective in MBRL is the expected cost over the parameter distribution of the dynamics model which reflects the epistemic uncertainty: 
\begin{equation}
    \overbar{C}(\boldsymbol{u}) \coloneq \mathbb{E}_{\boldsymbol{\theta}}\left[ C_{ \boldsymbol{\theta}}(\boldsymbol{u})\right] , \label{eq:meancostfunctional}
\end{equation}
where
\begin{align*}
      &&C_{ \boldsymbol{\theta}}(\boldsymbol{u}) &\coloneq 
      \int_{0}^{t_f} L\bigl(\boldsymbol{x}_{ \boldsymbol{\theta}}(\tau), \boldsymbol{u}(\tau)\bigr) \df\tau + \Phi \bigl(\boldsymbol{x}_{ \boldsymbol{\theta}}(t_f)\bigr),
\end{align*}
and $\mathbb{E}_{\boldsymbol{\theta}}$ is shorthand for $\mathbb{E}_{\boldsymbol{\theta} \sim p(\boldsymbol{\theta}\mid\mathcal{D})}$. This results in the optimal control problem:
\begin{align}
    \boldsymbol{u}^* (t) &\coloneq \argmin_{\boldsymbol{u}\in \mathcal{U}} \overbar{C}(\boldsymbol{u})  \\
    \text{s.t.} \quad \dot{\boldsymbol{x}}_{ \boldsymbol{\theta}}(t) &= \boldsymbol{f}_{ \boldsymbol{\theta}}\bigl(\boldsymbol{x}_{ \boldsymbol{\theta}}(t),\boldsymbol{u}(t), t\bigr),\quad \text{for } t \in [t_0, t_f] \enspace \nonumber\\ 
    \boldsymbol{x}_{ \boldsymbol{\theta}}(t_0) &= \boldsymbol{x}_0  . \nonumber   
\end{align}

\section{Probabilistic control via PMP}

\subsection{Pontryagin's maximum principle}
The mean cost objective $\overbar{C}(\boldsymbol{u})$ can be optimized via the \textit{direct} method or via the \textit{indirect} method. In the former, the control function is discretized and optimized as a real-valued vector, whereas in the latter, first the necessary optimality conditions for the continuous-time optimal control problem are described using PMP and then the control function is discretized. A key function is the Hamiltonian: 
\begin{equation}
    \mathcal{H}\bigl(\boldsymbol{x}, \boldsymbol{\lambda}, \boldsymbol{u},t \bigr) \coloneq L\bigl(\boldsymbol{x},\boldsymbol{u}\bigr) + \boldsymbol{\lambda}^{\top} \boldsymbol{f}\bigl(\boldsymbol{x},\boldsymbol{u},t\bigr), \nonumber \label{eq:Hamiltonian}
\end{equation}
where $\mathcal{H}\colon \mathcal{X} \times \R^{d_x} \times \mathcal{U} \times \R \to \R$ and $\boldsymbol{\lambda} \in \R^{d_x}$ are the co-states (or adjoints). For a given function $\boldsymbol{u}(t)$, the state and costate equations are given as:
\begin{align}
    \dot{\boldsymbol{x}}(t) &= \boldsymbol{f}\bigl(\boldsymbol{x}(t), \boldsymbol{u}(t),t \bigr), \label{eq:xdot_condition}\\
    \dot{\boldsymbol{\lambda}}(t) &= -\nabla_{\boldsymbol{x}} \mathcal{H} \bigl(\boldsymbol{x}(t),\boldsymbol{\lambda}(t),\boldsymbol{u}(t),t\bigr),\label{eq:lambdadot_condition} \\
     \boldsymbol{\lambda}(t_f) &= \nabla \Phi\bigl(\boldsymbol{x}(t_f)\bigr), \label{eq:transversality_condition}
\end{align}
for $t\in [t_0,t_f]$, where the notation $(\nabla_{y} f)_{i,j} \coloneq \partial f_j / {\partial y_i}$ denotes Jacobians, in this case $\nabla_{\boldsymbol{x}} \mathcal{H} = ( \frac{\partial \mathcal{H}}{\partial x_1}, \dots, \frac{\partial \mathcal{H}}{\partial x_{d_x}})^{\top}$ and $\nabla \Phi = ( \frac{\partial \Phi }{\partial x_1}, \dots, \frac{\partial \Phi}{\partial x_{d_x}})^{\top}$. Equation~\eqref{eq:transversality_condition} is commonly known as the \textit{transversality condition}, and provides a boundary condition for the co-states. 
If $\boldsymbol{u}^* (t)$ is an optimal solution to $C(\boldsymbol{u})$, the following necessary condition of optimality is derived via PMP:
\begin{equation}\label{eq:PMP_minimum_principle}
    \boldsymbol{u}^*(t) = \argmin_{\boldsymbol{u}}\mathcal{H}\left(\boldsymbol{x}^*(t),\boldsymbol{\lambda}^*(t),\boldsymbol{u},t\right),
\end{equation}
where $\boldsymbol{x}^*(t)$ and $\boldsymbol{\lambda}^*(t)$ satisfy \eqref{eq:xdot_condition} -- \eqref{eq:transversality_condition}. We refer to \eqref{eq:PMP_minimum_principle} as the \textit{minimum principle}. The indirect approach thus provides candidate solutions for the optimal control $\boldsymbol{u}^* (t)$ that satisfy the necessary optimality conditions for the cost functional. Next, we extend these conditions to the probabilistic setup.

\subsection{Probabilistic Hamiltonian}
\begin{figure}[t]
    \centering
    \includegraphics[width=1.\linewidth]{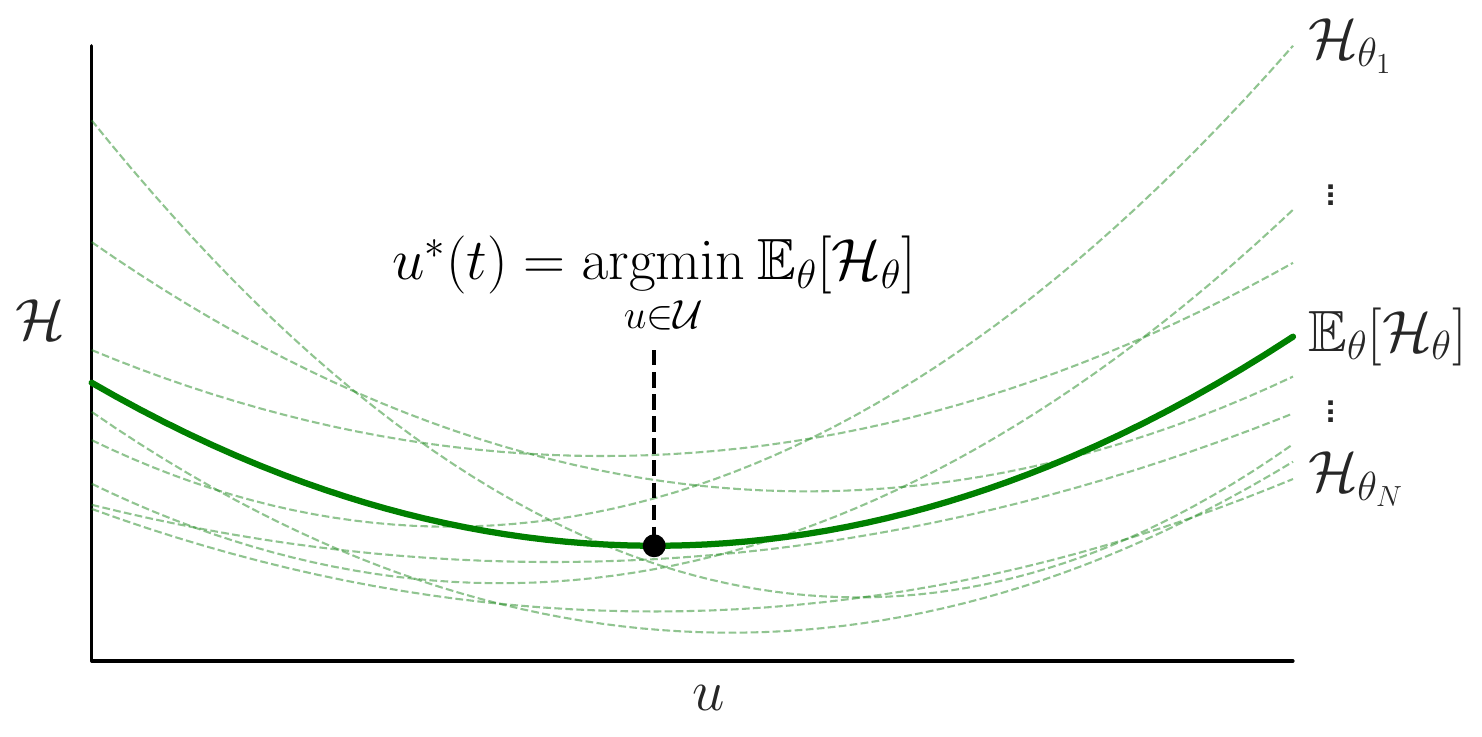}
    \caption{Illustration of the mean Hamiltonian minimization principle. At a given time $t$, each of the $N$ samples from a probabilistic dynamics model (indexed by $\boldsymbol{\theta}$) defines a Hamiltonian $\mathcal{H}_{\boldsymbol{\theta}} \coloneq \mathcal{H}_{ \boldsymbol{\theta}}\left(\boldsymbol{x}_{ \boldsymbol{\theta}}(t),\boldsymbol{\lambda}_{ \boldsymbol{\theta}}(t), \boldsymbol{u},t \right)$ as a function of the control $\boldsymbol{u}$ (dashed lines). The optimal control $\boldsymbol{u}^*$ minimizes the mean Hamiltonian, $\mathbb{E}_{\boldsymbol{\theta}}[\mathcal{H}_{\boldsymbol{\theta}}]$ (solid line), providing a control action that is optimal on average with respect to the posterior distribution over system dynamics.}
\label{fig:mean_hamiltonian_minimization}
\end{figure}

In the presence of uncertainty, the posterior distribution over system dynamics $p(\boldsymbol{\theta}\mid \mathcal{D})$ in \eqref{eq:BayesPosterior} implicitly defines a distribution over Hamiltonian functions $p(\mathcal{H}\mid \mathcal{D})$ through its dependence on the dynamics. Sampling a realization $\boldsymbol{\theta} \sim p(\boldsymbol{\theta}\mid \mathcal{D})$ yields a corresponding Hamiltonian:
\begin{align}
    \mathcal{H}_{ \boldsymbol{\theta}}(\boldsymbol{x}, \boldsymbol{\lambda}, \boldsymbol{u}, t) \coloneq L\left(\boldsymbol{x}, \boldsymbol{u} \right) +  \boldsymbol{\lambda}^{\top} \boldsymbol{f}_{ \boldsymbol{\theta}}\left(\boldsymbol{x},\boldsymbol{u},t\right).
\end{align}
which extends the classical Hamiltonian by incorporating the parameter-dependent dynamics $\boldsymbol{f}_{\boldsymbol{\theta}}$. 
Our goal is to find a control function $\boldsymbol{u}^* (t)$ that minimizes the mean cost functional $\overbar{C}(\boldsymbol{u})$. For a single realization $\boldsymbol{\theta}$, the state and co-state equations are:
\begin{align}
    \dot{\boldsymbol{x}}_{\boldsymbol{\theta}}(t) &= \boldsymbol{f}_{\boldsymbol{\theta}}\bigl(\boldsymbol{x}_{\boldsymbol{\theta}}(t), \boldsymbol{u}(t),t \bigr), \label{eq:xdot_conditiontheta}\\
    \dot{\boldsymbol{\lambda}}_{\boldsymbol{\theta}}(t) &= -\nabla_{\boldsymbol{x}} \mathcal{H}_{\boldsymbol{\theta}} \bigl(\boldsymbol{x}_{\boldsymbol{\theta}}(t),\boldsymbol{\lambda}_{\boldsymbol{\theta}}(t),\boldsymbol{u}(t),t\bigr),\label{eq:lambdadot_conditiontheta} \\
     \boldsymbol{\lambda}_{\boldsymbol{\theta}}(t_f) &= \nabla \Phi\bigl(\boldsymbol{x}_{\boldsymbol{\theta}}(t_f)\bigr). \label{eq:transversality_conditiontheta}
\end{align}
Here, $\boldsymbol{x}_{ \boldsymbol{\theta}}(t)$ and $\boldsymbol{\lambda}_{ \boldsymbol{\theta}}(t)$ are the state and costate trajectories for a give admissible control $\boldsymbol{u}(t)$ and realization $\boldsymbol{\theta}$ satisfying \eqref{eq:xdot_conditiontheta} -- \eqref{eq:transversality_conditiontheta}. The necessary optimality condition for a single realization $C_{ \boldsymbol{\theta}}(\boldsymbol{u})$ follows from the minimum principle:
\[
    \boldsymbol{u}_{ \boldsymbol{\theta}}^*(t) \coloneq \argmin_{\boldsymbol{u}} \mathcal{H}_{ \boldsymbol{\theta}}\left(\boldsymbol{x}_{\boldsymbol{\theta}}^* (t), \boldsymbol{\lambda}_{\boldsymbol{\theta}}^*(t), \boldsymbol{u},t\right),
\]
for $t\in [t_0,t_f]$, with $\boldsymbol{x}_{ \boldsymbol{\theta}}^*(t)$ and $\boldsymbol{\lambda}_{ \boldsymbol{\theta}}^*(t)$ satisfying \eqref{eq:xdot_conditiontheta} -- \eqref{eq:transversality_conditiontheta}. However, optimizing for a single $\boldsymbol{\theta}$ does not address the mean cost $\overbar{C}(\boldsymbol{u})$, which requires a control function that is optimal across the distribution of $\boldsymbol{\theta}$. To this end, we consider the \textit{mean minimum principle}:
\begin{equation} 
    \boldsymbol{u}^*(t) = \argmin_{\boldsymbol{u}}\mathbb{E}_{\boldsymbol{\theta} }\left[\mathcal{H}_{ \boldsymbol{\theta}}\left(\boldsymbol{x}_{ \boldsymbol{\theta}}(t),\boldsymbol{\lambda}_{ \boldsymbol{\theta}}(t), \boldsymbol{u},t \right)\right], \label{eq:meanhamiltonian}
\end{equation}
This condition yields a control function that is optimal in expectation, producing a family of state and co-state trajectories $\boldsymbol{x}_{ \boldsymbol{\theta}}(t)$ and $\boldsymbol{\lambda}_{ \boldsymbol{\theta}}(t)$ that satisfy \eqref{eq:xdot_conditiontheta} -- \eqref{eq:transversality_conditiontheta} for each $\boldsymbol{\theta}$. The mean minimum principle balances performance across all possible dynamics models weighted by their posterior probability, which is illustrated in Figure~\ref{fig:mean_hamiltonian_minimization}.

\subsection{Necessary optimality condition for the mean cost}
We now state the necessary condition for optimality for the mean cost $\overbar{C}(\boldsymbol{u})$. The general principle developed in ~\cite{bettiol2019averagecost} and~\cite{scagliotti2023optimal} requires the optimal control $\boldsymbol{u}^*(t)$ to minimize the mean Hamiltonian at each point in time over the (possibly bounded) set of admissible controls $\mathcal{U}$. Here, we restrict ourselves to the case of unconstrained control  setting, in which case this simplifies to a first-order condition on the gradient using a variational approach~\cite{bryson1975applied}.

\textit{Assumptions:} we assume that $\mathcal{U} = \mathbb{R}^{d_u}$, $\boldsymbol{f}_{\boldsymbol{\theta}}$ and $L$ are continuously differentiable w.r.t. $\boldsymbol{x}$ and $\boldsymbol{u}$, ensuring $\mathcal{H}_{\boldsymbol{\theta}}$ is differentiable for all $\boldsymbol{\theta} \in \Theta$, and \(\nabla_{\boldsymbol{u}} \mathcal{H}_{\boldsymbol{\theta}}\) satisfies standard regularity conditions for interchanging $\mathbb{E}_{\boldsymbol{\theta}}$ with differentiation and integration over $t \in [t_0, t_f]$.

\begin{theorem}[Necessary Mean Cost Optimality Condition] \label{thm:mean_pmp}
Let $\boldsymbol{u}^*(t)$ be an optimal control that minimizes the mean cost functional $\overbar{C}(\boldsymbol{u})$ over the set of admissible controls $\mathcal{U}$. Let the mean Hamiltonian be $\mathcal{\overbar{H}^*} \coloneq \mathbb{E}_{\boldsymbol{\theta}}[\mathcal{H}_{ \boldsymbol{\theta}}\bigl(\boldsymbol{x}_{ \boldsymbol{\theta}}(t),\boldsymbol{\lambda}_{ \boldsymbol{\theta}}(t), \boldsymbol{u}^* (t),t  \bigr) ]$, evaluated under $\boldsymbol{u}^* (t)$. A necessary condition for optimality is that:
\begin{equation}
    \nabla_{\boldsymbol{u}} \mathcal{\overbar{H}^*} = 0 \qquad \text{for all }  t \in [t_0, t_f]  .\label{eq:meanminimumprinciple}
\end{equation}
Since $\boldsymbol{u}^* (t)$ minimizes the mean cost, $\boldsymbol{x}_{ \boldsymbol{\theta}}(t)$ and $\lambda_{ \boldsymbol{\theta}}(t)$ are not necessarily optimal for each $\boldsymbol{\theta}$.
\end{theorem}

\begin{proof}
We consider the Gateaux derivative of the mean cost:
\[ 
    \delta \overbar{C}(\boldsymbol{u}) = \lim\limits_{\epsilon \to 0} \frac{\overbar{C}(\boldsymbol{u}+\epsilon \delta \boldsymbol{u})  -  \overbar{C}(\boldsymbol{u})}{\epsilon} ,
\]
where $\delta \boldsymbol{u}(t)$ is an arbitrary variation defined over $[t_0, t_f]$. The first variation can be obtained by isolating the first order terms around the optimal solution $\boldsymbol{u}^* (t)$:
\[
    \overbar{C}(\boldsymbol{u}^* + \epsilon \delta \boldsymbol{u}) - \overbar{C}(\boldsymbol{u}^*) \approx \delta \overbar{C}\big|_{\boldsymbol{u}^*(t)}(\delta \boldsymbol{u}) \epsilon ,
\]
where higher-order terms are neglected as $\epsilon \to 0$ . Since by definition $\boldsymbol{u}^* (t)$ is a minimum for $\overbar{C}$, its first variation must be zero. 
We first consider the deterministic case for a realization $\boldsymbol{\theta}$, where $\boldsymbol{u}_{ \boldsymbol{\theta}}^*(t)$ is optimal for $C_{ \boldsymbol{\theta}}(\boldsymbol{u})$. A well-known result in optimal control theory states that the first variation is given as \cite[Eq. (5.1.13)]{kirk2004optimal}:
\begin{align*}
    \delta C_{\boldsymbol{\theta}}\big|_{\boldsymbol{u}_{ \boldsymbol{\theta}}^*(t)}(\delta \boldsymbol{u}) = \int^{t_f}_{t_0} \bigl[ &\nabla_{\boldsymbol{u}} \mathcal{H}_{\boldsymbol{\theta}}^{*}\bigr] \delta \boldsymbol{u}(t) \df t ,
\end{align*}
where $\mathcal{H}_{ \boldsymbol{\theta}}^{*}\coloneqq\mathcal{H}_{ \boldsymbol{\theta}}\bigl( \boldsymbol{x}_{ \boldsymbol{\theta}}^*(t), \boldsymbol{\lambda}_{ \boldsymbol{\theta}}^*(t), \boldsymbol{u}_{ \boldsymbol{\theta}}^*(t),t\bigr)$.

By the fundamental lemma of calculus of variations, it follows that:
\[ 
\nabla_{\boldsymbol{u}} \mathcal{H}_{ \boldsymbol{\theta}}^*  = 0, \quad \text{for } t \in [t_0, t_f] .
\]
Using this result, we now turn to the mean cost and interchange the derivative and expectation:
\[
    \delta \overbar{C}(\boldsymbol{u}) = \delta \mathbb{E}_{\boldsymbol{\theta}} [C_{ \boldsymbol{\theta}}(\boldsymbol{u})] = \mathbb{E}_{\boldsymbol{\theta}} [\delta C_{ \boldsymbol{\theta}}(\boldsymbol{u})] .
\]
Using the earlier result of the first variation:
\begin{align*}
     &\mathbb{E}_{\boldsymbol{\theta}}[\delta C_{\boldsymbol{\theta}}\big|_{\boldsymbol{u}^*(t)}(\delta \boldsymbol{u})] \\
     &= \mathbb{E}_{\boldsymbol{\theta}} \bigl[ \int^{t_f}_{t_0} \nabla_{\boldsymbol{u}} \mathcal{H}_{ \boldsymbol{\theta}}\bigl(\boldsymbol{x}_{\boldsymbol{\theta}}(t),\boldsymbol{\lambda}_{ \boldsymbol{\theta}}(t), \boldsymbol{u}^*(t),t \bigr )   \delta \boldsymbol{u}(t) \df t \bigr]\\ 
     &= \int^{t_f}_{t_0} \nabla_{\boldsymbol{u}} \mathbb{E}_{\boldsymbol{\theta}} \bigl[\mathcal{H}_{ \boldsymbol{\theta}}\bigl(\boldsymbol{x}_{ \boldsymbol{\theta}}(t),\boldsymbol{\lambda}_{ \boldsymbol{\theta}}(t), \boldsymbol{u}^*(t),t  \bigr )\bigr] \delta \boldsymbol{u}(t) \df t  \\
     &= \int^{t_f}_{t_0}  \bigl[ \nabla_{\boldsymbol{u}} \mathcal{\overbar{H}^*}\bigr] \delta \boldsymbol{u}(t) \df t .
\end{align*}
By the fundamental lemma of the calculus of variations, it follows that:
\[
\nabla_{\boldsymbol{u}} \mathcal{\overbar{H}^*} = 0 \qquad \text{for all }  t \in [t_0, t_f] . \qedhere
\] 
\end{proof} This shows that the optimal control $\boldsymbol{u}^* (t)$ is only required to satisfy the optimality conditions in the expectation of $p(\boldsymbol{\theta} \mid \mathcal{D})$, extending the minimum principle to probabilistic dynamics models that capture epistemic uncertainty. 

\section{Numerical methods for probabilistic Hamiltonian}
\subsection{Trajectory sampling}
Computing the mean over Hamiltonians $\overbar{\mathcal{H}}$ in Eq. \eqref{eq:meanhamiltonian} requires estimating the uncertainty in state and costate trajectories. We adopt the trajectory sampling technique to capture this variability induced by the dynamics posterior~\cite{chua2018deep}. The mean Hamiltonian is approximated by aggregating the sampled Hamiltonian functions: 
\begin{equation}    
    \boldsymbol{u}^*(t) \approx \argmin_{\boldsymbol{u}}\frac{1}{M} \sum_{i=1}^{M} \mathcal{H}_{ \boldsymbol{\theta}_{i}}\left(\boldsymbol{x}_{ \boldsymbol{\theta}_{i}}(t), \boldsymbol{\lambda}_{\boldsymbol{\theta}_{i}}(t), \boldsymbol{u}, t\right),
\end{equation}
where $\boldsymbol{\theta}_{i} \sim p(\boldsymbol{\theta}\mid \mathcal{D})$ for $i = 1, \dots, M$, and each conditioned Hamiltonian is computed along the trajectories $\boldsymbol{x}_{\boldsymbol{\theta}_{i}}(t)$ and $\boldsymbol{\lambda}_{\boldsymbol{\theta}_{i}}(t)$, which comply with the PMP conditions. 
An alternative formulation is to consider the mean of the optimal controls $\overbar{\boldsymbol{u}}^*(t) = \mathbb{E}_{\boldsymbol{\theta}}[\boldsymbol{u}_{\boldsymbol{\theta}}^* (t)]$:
\begin{equation}
    \overbar{\boldsymbol{u}}^* (t) \approx \frac{1}{M} \sum^M_{i=1} \boldsymbol{u}_{\boldsymbol{\theta}_{i}}^* (t)  .
\end{equation}
By decoupling the optimization process over different realizations from the dynamics posterior, the mean optimal control approximation is only a valid necessary optimality condition when the Hamiltonian function is convex. However, the solutions can be computed in parallel, allowing for computationally efficient implementation since both approaches leverage Monte Carlo estimates of the state and co-state trajectories to efficiently evaluate the $M$ samples using GPU parallelization.

\subsection{Two-point boundary value problem} 
For each realization $\boldsymbol{\theta}_{i}$, the initial state $\boldsymbol{x}_{\boldsymbol{\theta}_{i}}(t_0) = \boldsymbol{x}_0$ and the final co-state $\boldsymbol{\lambda}_{\boldsymbol{\theta}_{i}}(t_f) = \nabla_{\boldsymbol{x}} \Phi \bigl( \boldsymbol{x}_{\boldsymbol{\theta}_{i}}(t_f)\bigr)$ are shared. This gives rise to a two-point boundary value problem (TPBVP) that satisfies the coupled differential equations $\dot{\boldsymbol{x}}_{\boldsymbol{\theta}_{i}}(t) = \boldsymbol{f}_{\boldsymbol{\theta}_{i}}\bigl( \boldsymbol{x}_{\boldsymbol{\theta}_{i}}(t), \boldsymbol{\lambda}_{\boldsymbol{\theta}_{i}}(t)\bigr)$ and $\dot{\boldsymbol{\lambda}}_{\boldsymbol{\theta}_{i}}(t)= - \nabla_{\boldsymbol{x}} \mathcal{H}_{\boldsymbol{\theta}_{i}}(\boldsymbol{x}_{\boldsymbol{\theta}_{i}}(t), \boldsymbol{\lambda}_{\boldsymbol{\theta}_{i}}(t), \boldsymbol{u}(t),t )$, where the initial value of the co-state $\boldsymbol{\lambda}_{\boldsymbol{\theta}_{i}}(t_0)$ is unknown. Due to the continuous-time formulation, numerical ODE solvers with adaptive step-size solvers can be leveraged to improve stability and control the numerical error.

We solve this using the \textit{forward shooting method}~\cite{diehl2011numerical}, which guesses an initial co-state $\boldsymbol{\lambda}_{\boldsymbol{\theta}_{i}} (t_0)$, integrates the states and co-states forward, and adjusts the guess based on the terminal errors $\boldsymbol{\lambda}_{\boldsymbol{\theta}_{i}}(t_f) - \nabla_{\boldsymbol{x}} \Phi \bigl( \boldsymbol{x}_{\boldsymbol{\theta}_{i}}(t_f)\bigr)$. Root-finding methods (e.g., Newton's method) or least-squares optimization (e.g., Levenberg-Marquardt) refine the guess iteratively.  For robustness with non-linear dynamics for long time horizons, we employ the \textit{multiple shooting} formulation~\cite{diehl2011numerical}. This method breaks down the long time horizon into multiple shorter intervals and enforces continuity conditions between these, to improve convergence and stability.

The mean Hamiltonian approach shares controls across dynamics function realizations, leading to $M S d_x$ decision variables, where $S$ is the number of shooting segments. In contrast, the mean optimal control variant decouples the optimization process of the sample realizations by only averaging \textit{after} the control function is optimized. This results in $M$ independent optimization problems with $S d_x$ decision variables, which can be solved in parallel. 

\subsection{Model predictive control}  
The TPBVP describes an open-loop trajectory optimization problem. To obtain a closed-loop control, principles from non-linear MPC can be leveraged by solving the trajectory optimization at every time point under a moving horizon of $t\in [t_0, t_0+H]$~\cite{diehl2011numerical}. The optimized control function is then applied until a new system measurement becomes available, and the optimization process is repeated. By using a finite prediction horizon, the control resulting from MPC is not necessarily optimal because it does not account for how future control actions will be re-optimized based on new measurements. Thus, this approach trades off theoretical optimality for computational tractability and the ability to react to real-time disturbances.

\section{Experimental Results}
We demonstrate our approach on three classical control problems, with all system parameters, control cost matrices and simulation parameters detailed in the Appendix, Table~\ref{tab:appendix_table}. For each task, we adopt a standard quadratic cost functional of the form \begin{align*}
&C(\boldsymbol{u}) = (\boldsymbol{x}(t_f) - \boldsymbol{x}^*)^{\top} Q_f (\boldsymbol{x}(t_f) - \boldsymbol{x}^*) \\ &+ \int_{t_0}^{t_f} \left( (\boldsymbol{x}(t)-\boldsymbol{x}^*)^{\top} Q (\boldsymbol{x}(t)-\boldsymbol{x}^*) + \boldsymbol{u}(t)^{\top} R \boldsymbol{u}(t) \right) dt \end{align*} where $Q$ and $Q_f$ are positive semi-definite state-cost matrices, $R$ is a positive definite control-cost matrix, and $\boldsymbol{x}^*$ is the target state.
The code implementing the experiments is available at https://github.com/DavidLeeftink/probabilistic-pontryagin-control. 

\subsection{Van der Pol stabilization}
We first evaluate our method on stabilizing a Van der Pol oscillator, a non-linear dynamical system commonly used in electrical circuit design and computational neuroscience~\cite{vanderpol1926relaxation}. We consider the following control formulation:
\begin{align*}
    \dot{x}_1 &= \mu \left(x_2 - \frac{1}{3}x_1^3 - x_1\right) , \\
    \dot{x}_2 &= -x_1 + u,     
\end{align*}
where $\mu = 1.5$. The goal is stabilization of the system at the unstable fixed point at the origin, under a quadratic cost function. This requires precise control actions to prevent repulsion towards the attracting limit cycle.   

We formulate an offline RL task where we generate a dataset by simulating the Van der Pol system for $n=25$ trajectories of length $T=10$. These simulations are driven by random periodic control inputs (Schroeder sweeps~\cite{schroeder1970synthesis}) and initialized from uniformly sampled states. In this offline setting, the system's underlying dynamics are assumed to be unknown. We then use this data to train a probabilistic dynamics model — an ensemble of neural ODEs — to capture the epistemic uncertainty. Finally, the learned model is used to construct a controller, allowing a direct comparison between the probabilistic Pontryagin approaches and direct optimization approaches of the mean cost.
 
To explicitly investigate the effect of epistemic uncertainty modeling, we compare our approach using both deterministic and probabilistic dynamics models trained on identical datasets.
We benchmark the mean Hamiltonian (PMP-Mean $\mathcal{H}$) and mean posterior (PMP-Mean $u^*$) formulations against direct optimization methods: Sequential Quadratic Programming (SQP), Broyden–Fletcher–Goldfarb–Shanno (BFGS)~\cite{nocedal1999numerical}, improved Cross-Entropy Method (iCEM)~\cite{rubinstein1999cross, pinneri2021sample}, and Adam~\cite{kingma2015adam}. We provide further experimental details in the Appendix, Table~\ref{tab:appendix_table}.
\begin{table}[t]
   \caption{Comparison of optimization methods for Van der Pol stabilization with true and learned dynamics (mean cost $\pm$ std. dev. over 15 repetitions). } 
   \label{tab:vanderpol_results}
   \centering
   \resizebox{\columnwidth}{!}{
   \begin{tabular}{l c c c}
   \toprule
   \textbf{Method} & \textbf{True Model} & \textbf{NODE} & \textbf{Prob. NODE} \\ 
   \midrule
   PMP-Mean $\mathcal{H}$ & \multirow{2}{*}{\textbf{9.85}}  &  \multirow{2}{*}{\textbf{12.71 $\pm$ 2.45}}  &  11.57 $\pm$ 1.73   \\ 
   PMP-Mean $u^*$&     &                     &   \textbf{10.59 $\pm$ 0.88}        \\ 
   iCEM& 10.08                  & 13.90 $\pm$ 2.80         & 14.06 $\pm$ 2.81           \\
   SQP& 10.04                   & 13.10 $\pm$ 2.52         & 13.69 $\pm$ 2.14        \\
    BFGS & 10.03                & 13.13 $\pm$ 2.59         & 13.67 $\pm$ 2.75       \\
   Adam    & 10.05              & 14.03 $\pm$ 2.66         &  14.27 $\pm$ 3.03    \\
   \bottomrule
   \end{tabular}
   }
\end{table}

Table \ref{tab:vanderpol_results} summarizes the performance in terms of accumulated cost over 15 repetitions. The indirect PMP approach achieves improved cost performance over its direct counterpart, under both the deterministic and probabilistic dynamics model. The lowest cost is achieved under the probabilistic Pontryagin approaches, where the mean optimal control approximation slightly outperforms the mean Hamiltonian approach, suggesting that explicitly modeling the epistemic uncertainty improves performance.

\subsection{Cart pole swing up}
\begin{table}[t]
   \caption{Comparison of optimization methods for the Cart pole control problem with true and learned dynamics (mean cost $\pm$ std. dev. over 15 repetitions). } 
   \label{tab:cartpole_results}
   \centering
   \resizebox{\columnwidth}{!}{
   \begin{tabular}{l c c c}
   \toprule
   \textbf{Method} & \textbf{True Model} & \textbf{NODE} & \textbf{Prob. NODE} \\ 
   \midrule
   
   PMP-Mean $\mathcal{H}$& \multirow{2}{*}{12.57}  &  \multirow{2}{*}{112.22 $\pm$ 126.99}  & \textbf{36.78 $\pm$ 31.45}   \\ 
   PMP-Mean  $u^*$ &     &                     & 64.44 $\pm$ 77.95           \\ 
   iCEM& 7.23                  & $\:\:$\textbf{87.55 $\pm$ 109.56}         & 109.51 $\pm$ 183.34            \\
   SQP& 12.57                   & 111.52 $\pm$ 110.37         & $\:\:$77.98 $\pm$ 107.10            \\
    BFGS & 12.57             & 147.95 $\pm$ 168.82         & $\:\:$81.78 $\pm$ 132.58            \\
   Adam    & 12.89              & 251.31 $\pm$ 270.43        &  311.77 $\pm$ 349.75           \\
   \bottomrule
   \end{tabular}
   }
\end{table}
Next, we consider a cart pole swing-up task, where the objective is to balance the pendulum upright from an initial down position under quadratic cost~\cite{kelly2017introduction}:
\begin{align*}
    \ddot{x} &= \frac{l m \sin(\theta) \dot{\theta}^2 + u + m\, g \cos(\theta) \sin(\theta)}{M + m (1 - \cos^2(\theta))} ,\\
    \ddot{\theta} &= -\frac{l m  \cos (\theta) \sin(\theta) \dot{\theta}^2 +  u\cos\theta + (M + m)\, g \sin(\theta)}{l M + l m (1 - \cos^2(\theta))} .
\end{align*}
Here, $x$ represents the horizontal position of the cart, $\theta$ is the angle of the pole (with $\theta=\pi$ corresponding to the upright, balanced position), and the control input $u$ is the horizontal force applied to the cart. The physical parameters are the pole's half-length $l$, the pole's mass $m$, the cart's mass $M$, and the acceleration due to gravity $g$. 

We design an offline RL task in similar fashion to the Van der Pol experiment, where the dynamics of the cart pole are assumed to be unknown and have to be learned from observed  trajectories. We generate $n=50$ trajectories of length $T=5$ using random controls and uniformly sampled initial conditions, which serve as data. This results in sparse observations near the goal state. We compare the performance of the previously mentioned trajectory optimization methods on deterministic and probabilistic neural ODEs. Further experimental details are described in the Appendix, Table~\ref{tab:appendix_table}.

\begin{figure}[t]
    \centering
    \includegraphics[width=\columnwidth]{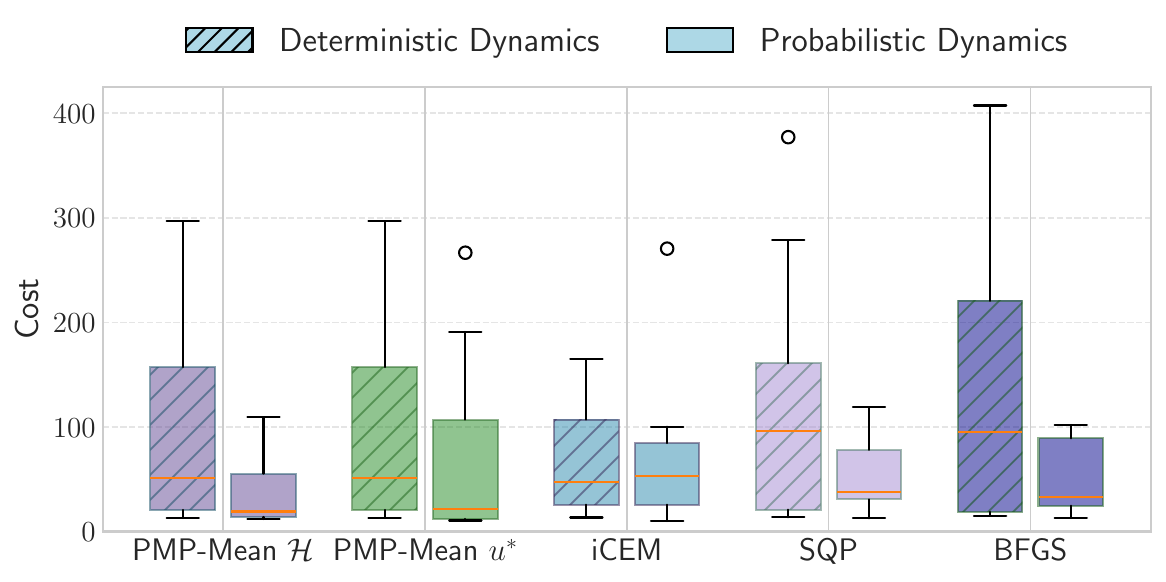}
    \caption{Comparison of optimization methods for the cart pole control task with deterministic and probabilistic dynamics models. The boxplot shows the distribution of cost over 15 datasets, including the median, interquartile range, and outliers. Due to its high cost average, the Adam optimizer is not visualized.}
    \label{fig:cartpole-boxplot}
\end{figure}

Table \ref{tab:cartpole_results} and Figure \ref{fig:cartpole-boxplot} present the trial costs averaged over 15 repetitions, comparing deterministic and probabilistic dynamics models across various optimization methods. Under limited available data, adopting the probabilistic dynamics model improves performance for most optimizers, emphasizing the role of capturing epistemic uncertainty in learning-based control tasks. The mean Hamiltonian formulation achieves the lowest mean and median trial costs out of the considered optimization methods, outperforming direct methods. The mean posterior approximation, theoretically suited only for convex Hamiltonians, performs worse than its counterpart with higher variability and average cost. 

\subsection{Duffing oscillator stabilization}
Lastly, we consider an online RL study involving the Duffing oscillator, a non-linear system modeling a mass in a double well potential, which can be regarded as a model of a steel beam that is deflected towards two magnets~\cite{moon1979magnetoelastic}:
\begin{align*}
    \dot{x}_1 &= x_2 , \\
    \dot{x}_2 &= -\delta x_2 -\alpha x_1 -\beta x_1^3 + \gamma u.
\end{align*}
Under periodic forcing, chaotic behavior can be observed. In our configuration, the system has two stable attractors, separated by an unstable fixed point at the origin. The objective is to drive the system to the unstable fixed point. 

In the online RL setting, one starts from a single random-policy trial, and iteratively trains the probabilistic dynamics model, updates the controller, and augments the dataset after each trial. We compare the mean Hamiltonian and mean posterior methods to the state-of-the-art PETS algorithm~\cite{chua2018deep}, which combines probabilistic dynamics models with MPC and the (i)CEM optimizer. We adapt PETS to the continuous-time setting using probabilistic neural ODEs as dynamics models. 
Further experimental details are described in the Appendix, Table~\ref{tab:appendix_table}.

\begin{figure}
    \centering
    \includegraphics[width=\columnwidth]{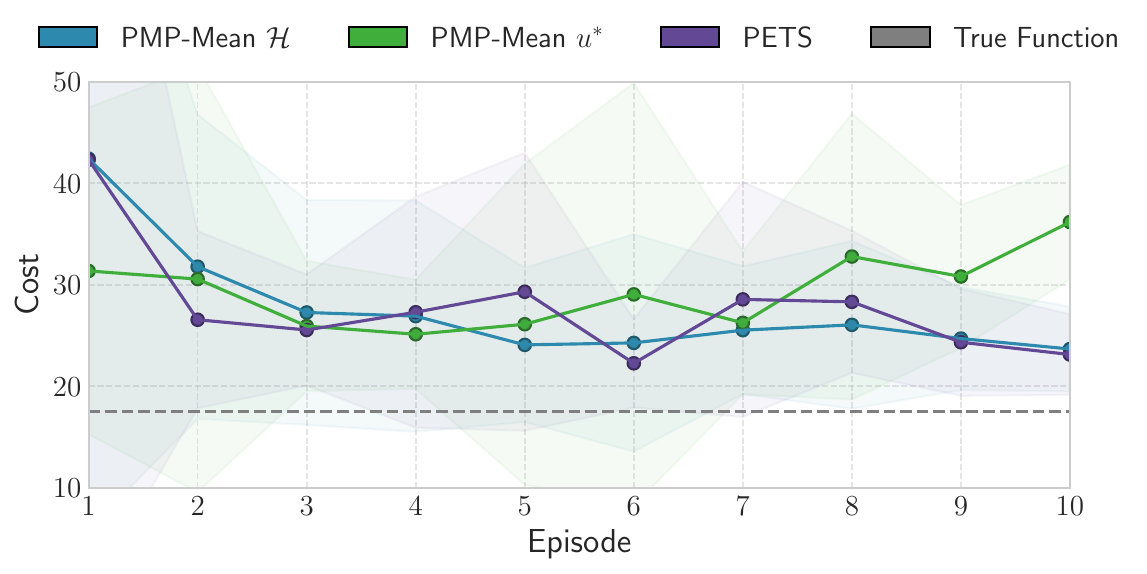}
    \caption{Comparison of online model-based RL methods for the Duffing oscillator stabilization task (cost $\pm$ std. dev. over 10 independent repetitions).}
    \label{fig:RL_duffing}
\end{figure}
Figure \ref{fig:RL_duffing} presents the average trial costs over 10 sequential iterations (averaged over 10 independent repetitions), with the dashed reference line indicating performance under known dynamics. Both PETS and the mean Hamiltonian approach converge to similar final costs, although PETS shows slight cost oscillations. The mean posterior approximation initially performs better but later becomes unstable, likely due to the inadequate aggregation of independently optimized control functions in the presence of multiple attractors near the goal state.

\section{Conclusion and discussion}
In this work, we have introduced a novel framework for MBRL that addresses the critical challenge of planning under epistemic uncertainty. Our approach translates a probabilistic interpretation of PMP, established in optimal control theory \cite{bettiol2019averagecost, scagliotti2023optimal}, into a practical algorithm for systems with learned dynamics. The core of our method is the principle of minimizing the mean Hamiltonian, which we compute over an ensemble of models representing the posterior distribution over the system dynamics that arises from limited data.

We developed a scalable numerical method, combining forward multiple shooting with moving horizon optimization, to make this principled approach tractable for deep, probabilistic models. A key feature of our framework is that it optimizes trajectories in continuous time—naturally aligning with the physics of many real-world systems—while still only requiring standard discrete-time observations for model learning. While we adopted neural ODE ensembles as a probabilistic dynamics model, the proposed principle generalizes to other probabilistic dynamics models, including Bayesian neural ODEs or systems with known functional forms but uncertain parameters.

Our experiments demonstrate the practical effectiveness of our approach in offline RL tasks under limited data availability, including the Van der Pol oscillator and a cart pole system. Comparisons against direct trajectory optimization methods, such as SQP and (i)CEM, show that our proposed mean Hamiltonian formulation achieves reduced trial costs. Our results further underline the advantage of explicitly modeling the epistemic uncertainty: under identical neural architectures, the probabilistic ensemble model consistently outperforms its deterministic counterpart, indicating that capturing epistemic uncertainty can lead to improved control performance. Additionally, in online RL settings, such as the stabilization of a Duffing oscillator, the proposed approach achieves competitive trial cost and data efficiency, comparable to the state-of-the-art MBRL method PETS. 

Although these results are promising, computational challenges arise in higher-dimensional systems or large ensemble sizes, due to solving the co-state equations via a root-finding task with cubic complexity. However, the sparse Jacobian structure and parallelizable computations of the sampled dynamics provide avenues for future algorithmic improvements or GPU acceleration. Alternatively, the forward-backward iterative approach can be adopted, which avoids the cubic complexity but may require additional iterations until convergence~\cite{mitter1966successive}. 

Future theoretical work could explore extending the optimality conditions to problems with state constraints, which commonly arises in safety critical application. One possible avenue to realize this is by incorporating safety constraints in the Hamiltonian via additional co-state variables. From a practical perspective, our probabilistic PMP approach is especially suited for applications requiring precise control under uncertainty on slower timescales, such as personalized treatment planning in healthcare or process control.


\section*{ACKNOWLEDGMENTS}
We would like to thank Dr. Yuzhen Qin for his valuable feedback and insightful comments during the review of this manuscript. We are also grateful to the anonymous reviewers for their constructive suggestions, which helped to significantly improve the final version of the paper. 

\addtolength{\textheight}{-2.5cm}   
\section*{APPENDIX}\label{app:experiment_details}
\begin{table*}[t] 
\renewcommand{\arraystretch}{1.6}
\caption{System parameters, control cost matrices and simulation parameters used in the experiments.}
\label{tab:appendix_table}
\centering
\resizebox{\textwidth}{!}{%
\begin{tabular}{|c|c|c|c|c|c|c|c|c|c|c|c|c|}
\hline
\textbf{System} 
& \textbf{Parameters} 
& \(Q\) 
& \(Q_f\) 
& \(R\) 
& \(\sigma\) 
& \(\mathcal{U}\) 
& \(x_0\) 
& \(x^*\) 
& \(t_0\) 
& \(t_f\) 
& \(\Delta t\) (MPC) 
& \(H\) 
\\
\hline
\textbf{Van der Pol} 
&
$\mu = 1.5 $
& \(\mathbb{I}\) 
& \(\mathbb{I}\) 
& \(0.5\) 
& \(0.01\) 
& \([-2, 2]\) 
& \((1, 1)^{\top}\) 
& \((0, 0)^{\top}\) 
& \(0\) 
& \(10\) 
& \(0.05\) 
& \(3\) 
\\
\hline
\textbf{Cart pole} 
& 
\parbox{2.5cm}{ \begin{align*}
    l &= m = M = 1 \\ 
    g &= 9.81
\end{align*}
} 
&  \(\mathbb{I}(1, 1, 0.1, 0.1)^{\top}\) 
&  \(\mathbb{I}(1, 5, 1, 1)^{\top}\) 
& \(0.05\) 
& \(0\) 
& \([-20, 20]\) 
& \parbox{1.5cm}{%
\((0, 0, 0, 0)^{\top}\)} 
& \((1, \pi, 0, 0)^{\top}\) 
& \(0\) 
& \(5\) 
& \(0.02\) 
& \(1\) 
\\
\hline
\textbf{Duffing} 
& 
\parbox{1.3cm}{ \begin{align*}
    \alpha &= -1, \quad \beta = 2\\
    \delta &= 0.2, \quad \gamma = 1 
    \end{align*}
} 
& \(5 \cdot \mathbb{I}\) 
& \( 5 \cdot \mathbb{I}\) 
& \(1\) 
& \(0.01\) 
& \([-2, 2]\) 
& \((1.5, 1)^{\top}\) 
& \((0, 0)^{\top}\) 
& \(0\) 
& \(5\) 
& \(0.05\) 
& \(2\) 
\\
\hline
\end{tabular}
}
\end{table*}
\subsection{Experimental details}

For all experiments, cubic interpolation is used to construct a continuous-time control function from a discretized control vector, and a 5th-order Dormand-Prince solver is used to solve predicted states (and co-states for PMP)~\cite{dormand1980family} for the learned dynamics model. An overview of all the system parameters, control cost matrices and simulation parameters used in the experiments is presented in Table \ref{tab:appendix_table}. 

The following settings were used for the respective experiments:
\paragraph{Van der Pol} Following the approach described in~\cite{chua2018deep}, we consider an ensemble size of $M=5$ following common practice in model-based RL, with each neural ODE consisting of two hidden layers with 32 units, resulting in parameters $\boldsymbol{\theta} \in \R^{1248}$. Our indirect methods use $S=4$ shooting segments. All trajectory optimization methods employ moving horizon optimization (horizon $H=3$) and 15 iterations per timestep. 
\paragraph{Cart pole} The dynamics model architecture is consistent with the previous Van der Pol setup, using ensemble neural ODEs ($\boldsymbol{\theta} \in \R^{1376}$) with identical control interpolation and ODE solver. Here, we use a shorter MPC horizon of $H=1$ with 25 iterations per time step, and two shooting segments ($S=2$) to ensure computational efficiency. 
\paragraph{Duffing} We adopt an ensemble of neural ODEs with $M=5$ as a probabilistic dynamics model, consistent with the previous van der Pol setup ($\boldsymbol{\theta} \in \R^{1248}$) with identical control interpolation and ODE solver. We employ MPC with horizon $H=2$ and adjusted integration steps suitable for the Duffing oscillator's faster time scale, and use $S=5$ shooting segments for the indirect methods.


\bibliography{references_compact}  

\end{document}